\documentclass{article}

\PassOptionsToPackage{numbers, compress}{natbib}

\usepackage[final]{nips_2018}

\usepackage[utf8]{inputenc} 
\usepackage[T1]{fontenc}    
\usepackage{hyperref}       
\usepackage{url}            
\usepackage{booktabs}       
\usepackage{amsfonts}       
\usepackage{nicefrac}       
\usepackage{microtype}      
\usepackage{caption}
\usepackage[table,xcdraw]{xcolor}
\definecolor{myblue}{rgb}{0 0 0.4}
\usepackage{multirow}
\usepackage{graphicx}
\usepackage{verbatim} 
\usepackage{amsfonts}
\hypersetup{
    colorlinks,
    citecolor=myblue,
    filecolor=myblue,
    linkcolor=myblue,
    urlcolor=myblue
}

\title{PaccMann: Prediction of anticancer compound sensitivity with multi-modal attention-based neural networks}

\author{
  Ali Oskooei\thanks{Shared first-authorship. \texttt{\{osk,jab,tte\}@zurich.ibm.com}}\hspace{1.1mm}\textsuperscript{1}, Jannis Born\footnotemark[1]\hspace{1.1mm}\textsuperscript{1,2,3}, Matteo Manica\footnotemark[1]\hspace{1.1mm}\textsuperscript{1,3}, \\ \textbf{Vigneshwari Subramanian\textsuperscript{4}}, \textbf{Julio S\'aez-Rodr\'iguez\textsuperscript{4,5}}, \textbf{Mar\'ia Rodr\'iguez Mart\'inez\textsuperscript{1}} \\
\textsuperscript{1}IBM Research, Z\"urich, Switzerland, \textsuperscript{2}University of Z\"urich, \textsuperscript{3}ETH Z\"urich,\\ \textsuperscript{4}RWTH Aachen University, Aachen, Germany, \textsuperscript{5}Heidelberg University, Heidelberg, Germany\\
}
 
\begin{document}

\maketitle

\begin{abstract}
  We present a novel approach for the prediction of anticancer compound sensitivity by means of multi-modal attention-based neural networks (PaccMann). In our approach, we integrate three key pillars of drug sensitivity, namely, the molecular structure of compounds, transcriptomic profiles of cancer cells as well as prior knowledge about interactions among proteins within cells. Our models ingest a drug-cell pair consisting of SMILES encoding of a compound and the gene expression profile of a cancer cell and predicts an IC50 sensitivity value. Gene expression profiles are encoded using an attention-based encoding mechanism that assigns high weights to the most informative genes. We present and study three encoders for SMILES string of compounds: 1) bidirectional recurrent 2) convolutional 3) attention-based encoders. We compare our devised models against a baseline model that ingests engineered fingerprints to represent the molecular structure. We demonstrate that using our attention-based encoders, we can surpass the baseline model. The use of attention-based encoders enhance interpretability and enable us to identify genes, bonds and atoms that were used by the network to make a prediction. 
\end{abstract}

\section{Introduction}
\label{seq:intro}
Invention of novel compounds with a desired efficacy and improving existing therapies are key bottlenecks in the pharmaceutical industry, which fuel the largest R\&D business spending of any industry and account for 19\% of the total R\&D spending worldwide~\cite{petrova2014innovation, goh2017smiles2vec}.Anticancer compounds, in particular, take the lion’s share of drug discovery R\&D efforts, with over 34\% of all drugs in the global R\&D pipeline in 2018 (5,212 of 15,267 drugs)~\citep{lloyd2017pharma}.
Despite enormous scientific and technological advances in recent years, serendipity still plays a major role in anticancer drug discovery~\citep{hargrave2012serendipity} without a systematic way to accumulate and leverage years of R\&D to achieve higher success rates in drug discovery.

On the other hand, there is strong evidence that the response to anticancer therapy is highly dependent on the tumor genomic and transcriptomic makeup, resulting in heterogeneity in patient clinical response to anticancer drugs~\cite{garnett2012systematic, yang2012genomics}.
This varied clinical response has led to the promise of personalized (or precision) medicine in cancer, where molecular biomarkers, e.g., the expression of specific genes, obtained from a patient’s tumor profiling may be used to choose a personalized therapy.
These challenges highlight a need across both pharmaceutical and health care industries for multi-modal quantitative methods that can jointly exploit disparate sources of knowledge with the goal of characterizing the link between the molecular structure of compounds, the genetic and epigenetic alterations of the biological samples and drug response. There have been a plethora of works focused on prediction of drug sensitivity in cancer cells~\cite{garnett2012systematic, costello2014community, geeleher2016cancer}, however, the majority of them have focused on the analysis of unimodal datasets such as genomic or transcriptomic profiles of cancer cells. A few models have proposed the integration of two different data modalities, e.g., genomic features and chemical descriptors~\cite{menden2013machine}. 

Chemical descriptors and fingerprints are engineered features that describe the chemical properties and structure of a compound. Chemical descriptors have been extensively used in the  chemical and biological  engineering domain to develop quantitative structure-activity relationship (QSAR) and quantitative structure–property relationship (QSPR) models of compounds~\citep{karelson1996quantum}. Similarly,  molecular fingerprints (FP) have been extensively applied to drug discovery, virtual screening and compound similarity search~\cite{cereto2015molecular}. Even though chemical descriptors and fingerprints have found many applications in combination with machine learning algorithms, the usage of engineered features limit the learning ability of such algorithms and may prove problematic in applications where these features are not relevant or informative~\cite{goh2017smiles2vec}. 

\begin{figure}[!htb]
\centering
\includegraphics[width=.9\linewidth]{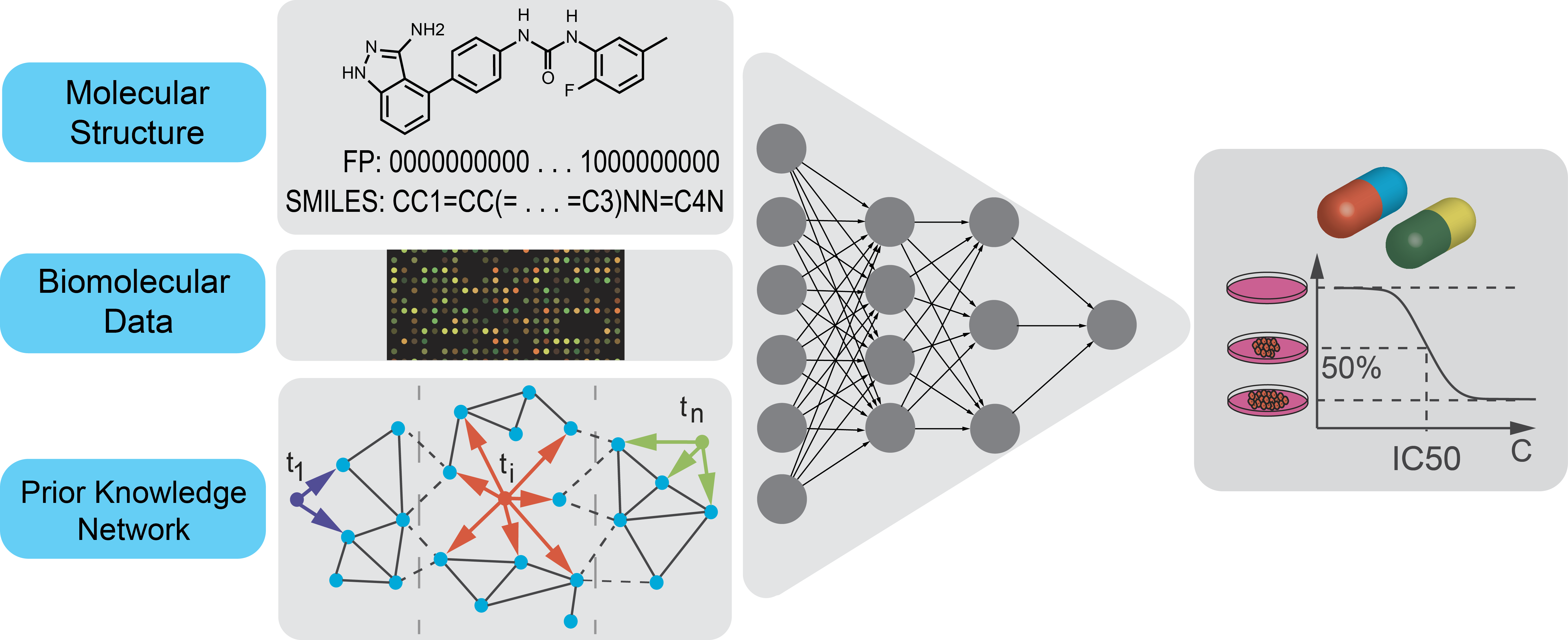}%
\caption{
\textbf{Multi-modal prediction of IC50 drug sensitivity.}
Three key data modalities that influence anticancer drug sensitivity: biomolecular measurements of cancer cells (gene expression, copy number alteration etc.), a network of known  interactions between the biomolecular entities and the chemical structure of the anticancer compounds. 
}
\label{fig:intro}
\end{figure}

With the  rise of deep learning methods and their proven ability to learn the most informative features from raw data, it seems imperative to approach chemical problems from a similar standpoint.
Recently, methods borrowed from neural language models~\cite{bahdanau2014neural} have been used in combination with SMILES string encoding of molecules to successfully predict chemical properties of molecules~\cite{goh2017smiles2vec, chen2018rise} or products of a chemical reaction~\cite{schwaller2018found}.
To the best of our knowledge there have not been any multi-modal deep learning solutions for anticancer drug sensitivity prediction that combine molecular structure of compounds, genetic profile of cells and prior knowledge of protein interactions. It is timely that utility of such a model be systematically explored and demonstrated. We address this gap in the current work. Specifically, we explore neural feature learning from raw SMILES encoding of chemicals (as opposed to engineered features) in the context of drug sensitivity in cancer cells (see \autoref{fig:intro}).
We validate our models against a baseline feedforward model based on the Morgan (Circular) chemical fingerprints~\cite{rogers_extended-connectivity_2010}. 
In doing so, we demonstrate that the direct analysis of raw SMILES encodings through an attention-based architecture can surpass the performance of models based on fingerprints. This is highly desirable, as SMILES are accessible and intuitively more  interpretable.

\section{Methods}
\label{seq:methods}

\subsection{Data}
\label{sseq:data}
Throughout this work, we employed the gene expression and drug IC50 data publicly available as part of the Genomics of Drug Sensitivity in Cancer (GDSC) database~\cite{garnett2012systematic, yang2012genomics}. 
The dataset includes the results of screening on  more than a thousand genetically profiled human pan-cancer cell lines with a wide range of anticancer compounds. 
The screened compounds include chemotherapeutic drugs as well as targeted therapeutics from various sources~\cite{yang2012genomics}.
We based our models on gene expression data, as it has been shown to be more predictive of drug sensitivity than genomics (i.e., copy number variation and mutations) or epigenomics (i.e., methylation) data~\cite{menden2016, oskooei2018network,costello2014community}.
STRING protein-protein interaction (PPI) network~\cite{szklarczyk2014string} was used to incorporate intracellular interactions in our model.
Within the GDSC database, 221 compounds were targeted therapies,  which are the focus of our work.
Of 221 compounds, the molecular structures of 208 compounds were publicly available. For the 208 compounds with publicly available molecular structure, we collected canonical SMILES and acquired Morgan fingerprints using \texttt{RDKit}\footnote{\url{https://github.com/rdkit/rdkit}}.
Exploiting the property that most molecules have multiple valid SMILES strings, we adopted a data augmentation strategy~\cite{bjerrum2017smiles} and represented each anticancer compound with 32 SMILES strings.
This results in a dataset consisting of 6,556,160 drug-cell pairs.

\subsection{Network propagation}
\label{sseq:propagation}
Network propagation was employed to reduce the dimensionality of the cell lines' transcriptomic profile,  consisting of ~16,000 genes, to a smaller subset containing the most informative genes~\cite{oskooei2018network}.
We identify the most relevant genes by employing a weighting and network propagation scheme: we first assign a high weight ($W=1$) to the reported drug target genes while assigning a very small positive weight ($\varepsilon=1\mathrm{e}{-5}$) to all other genes.
Thereafter, the initialized weights are propagated over the STRING protein-protein-interaction (PPI) network~\cite{szklarczyk2014string}, a comprehensive PPI database including interactions from multiple data sources.
This process is meant to integrate prior knowledge about molecular interactions into our weighting scheme, and simulates the propagation of perturbations within the cell following the drug administration.
The network smoothing of the weights can be described as a random walk, starting from the initial drug target weights, throughout the network.
More specifically, let us denote the initial weights as $W_{0}$ and the string network as $S = (P, E, A)$, where $P$ are the protein vertices of the network, $E$ are the edges between the proteins and $A$ is the weighted adjacency matrix, a matrix that indicates the level of confidence about the existence of a certain interaction.
The smoothed weights are determined from an iterative solution of the propagation function~\cite{oskooei2018network}:
\begin{equation}
\label{eq:propagation}
W_{t+1} =\alpha W_{t}A' + (1-\alpha)W_{0} \hspace{1cm} \textmd{where} \hspace{1cm} A' = D^{-\frac{1}{2}}AD^{-\frac{1}{2}}
\end{equation}
where $D$ is the degree matrix and  
$A'$ is the normalized adjacency matrix.
The diffusion tuning parameter, $\alpha$ ($0\leq\alpha\leq1$), defines how far the prior knowledge weights can diffuse through the network.
In this work, we used $\alpha=0.7$, as recommended in the literature for the STRING network~\cite{hofree2013network}.
Adopting a convergence rule of $e=(W_{t+1}-W_t) < 1 \mathrm{e}{-6}$, we solved  ~\autoref{eq:propagation} iteratively for each drug and used the resultant weights distribution to determine the top $20$ highly ranked genes for each drug.
We then pooled the top 20 genes for all 208 compounds, resulting in a subset of 2,128 genes. This subset containing the most informative genes was then used to profile each cell line in the dataset before it was fed into our models.

\subsection{Model architectures}
\begin{figure}[!htb]
\centering
\includegraphics[width=.7\linewidth]{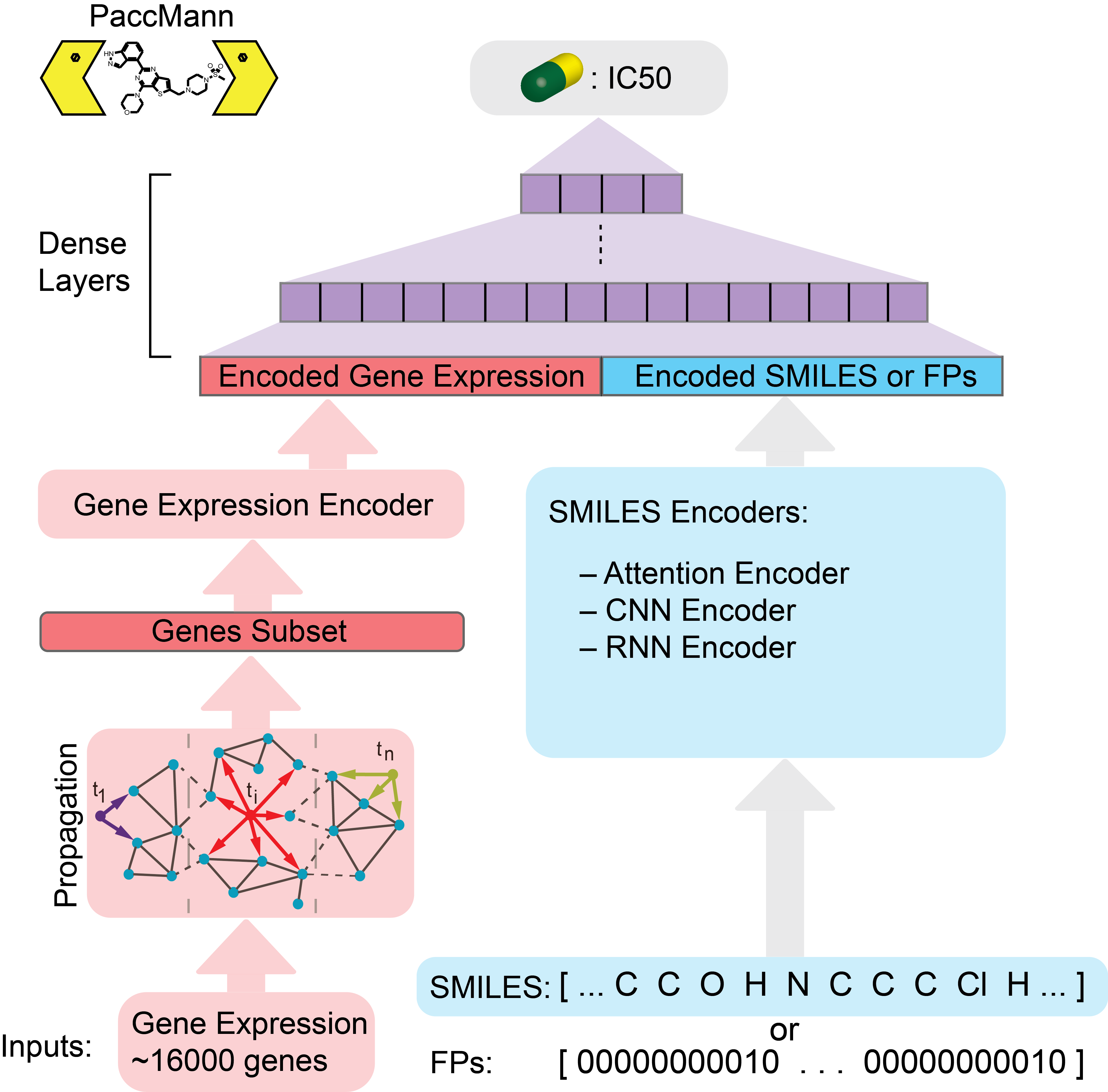}%
\caption{\textbf{The end-to-end architecture of PaccMann.} PaccMann ingests a cell-compound pair and makes a IC50 drug sensitivity prediction for the pair. Cells are represented by the gene expression values of a subset of 2,128 genes, selected for having the highest weights following the network propagation. Compound structures are represented either as 512 bit fingerprint or in the SMILES formats. 
The gene-vector is fed into an attention-based encoder that assigns higher weights to the most informative genes. Fingerprints are used in combination with the gene expression subset in a dense baseline model. SMILES encoding of compounds is employed by an array of encoders that are combined with a representation of gene expression to obtain a drug sensitivity prediction.}
\label{fig:architectures}
\end{figure}

The general architecture of our model is shown in \autoref{fig:architectures}.
To predict drug sensitivity we adopted two architecure configurations to ingest the different molecular structure encodings:
\\\textbf{Fingerprint models.}
We selected fingerprint-based models as the baseline comparison for all the considered model configurations.
Fingerprint representations of molecular structure have been proven to be a highly informative, and can therefore  serve as a reliable reference~\cite{unterthiner2014deep, ramsundar2015massively}.
Our baseline model is a 6-layered DNN with [512, 256, 128, 64, 32, 16] units and a sigmoidal activation designed by optimizing hyperparameters following a cross-validation scheme (see \autoref{sseq:data_split}).
The genetic profiles are filtered using the network propagation process described in \autoref{sseq:propagation}, and ingested in conjunction with 512-bit Morgan fingerprints as the encoding of the compound structure.
\\\textbf{SMILES models.}
These models ingest the filtered gene expression profiles and the SMILES text encodings for the  structure of the compounds.
To encode the SMILES, we employed 5 models falling into 3 general categories: recurrent, convolutional and pure attention-based encoders (see \autoref{fig:architectures}).

\begin{figure}[!htb]
\centering
\includegraphics[width=.7\linewidth]{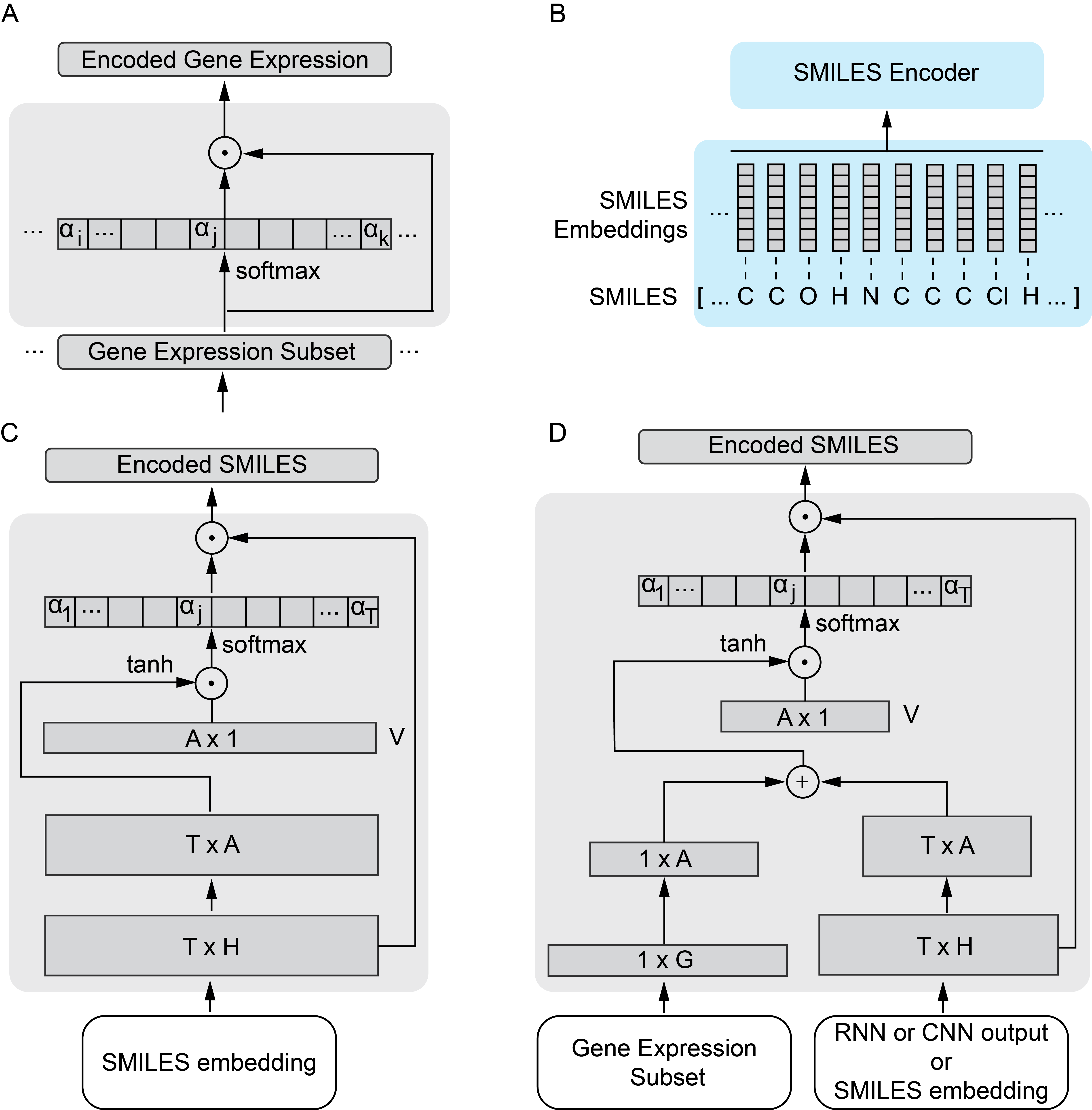}%
\caption{\textbf{The architecture of various encoders used in PaccMann.} A) An attention-based gene expression encoder generates attention weights that are in turn applied to the input gene subset via a dot product. B) An embedding layer transforms raw SMILES strings into a sequence of vectors in an embedding space. C) A self-attention SMILES encoder ingests the SMILES embeddings and computes attention weights,  $\alpha_{i}$, that is used to combine the inputs into a single vector of hidden dimensionality $H$. D) SMILES encoder with an attention process similar to Bahdanau’s, where the target hidden states are replaced by a context vector representing the gene expression subset~\cite{bahdanau2014neural}. The model learns to give higher weights to SMILES tokens that have the most influence given the gene expression of the cell, unlike the attention process described in C), where weights are determined based only on the SMILES.}
\label{fig:attention_mechanism}
\end{figure}

The genetic subset selected through network propagation is further processed using a gene attention encoder (GAE), see \autoref{fig:attention_mechanism}\textcolor{myblue}{A}. This architecture produces weights for each of the genes and applies these weights to the genes in a dot product, ensuring most informative genes receive a higher weight.
In addition, the resulting gene weights render the model more interpretable, as it identifies the  genes that drive the sensitivity prediction in each cell line.
To ensure charged or multi-character atoms (e.g., \texttt{Cl} or \texttt{Br}) were added as distinct tokens to the dictionary, the SMILES sequences were tokenized using a regular expression~\cite{schwaller2018found}.
The resulting atomic sequences of length $T$ are represented as $E=\{e_1,..., e_T\}$, with learned embedding vectors $e_i$ of length $H$ for each dictionary token (see \autoref{fig:attention_mechanism}\textcolor{myblue}{B}).

To investigate which model architecture best exploits the drug compound information, we explored various SMILES encoders, starting with a 2-layered bidirectional recurrent neural network (bRNN) that utilized gated recurrent units (GRU)~\cite{cho2014learning}. Next, we employed a stacked convolutional encoder (SCNN) with 4 layers and a sigmoidal activation function. Whilst 2D convolutions in the first layer collapsed the embedding hidden dimensionality, subsequent 1D convolutions extracted incrementally increasing long-range interactions between different parts of the molecule. 

As previously mentioned, we employed encoders that leveraged attention mechanisms. Interpretability is paramount in medicinal and machine learning applications to the health domain~\cite{koprowski2018machine}.  As such, attention in particular, is central in our models as it enables us to explain and interpret the observed results in the context of underlying biological and chemical processes. As such, we explored different attention-based encoders in our models. The first attention configuration we adopted is a self-attention (SA) mechanism originally developed for document classification \cite{yang2016hierarchical} and here adapted to SMILES representations (see \autoref{fig:attention_mechanism}\textcolor{myblue}{C}).
The attention weights $\alpha_{i}$ were computed as:
\begin{equation}
\label{eq:seq_att}
\alpha_{i} = \frac{\exp(u_{i})}{\sum_{j}^{T} \exp(u_{j}) } \hspace{1cm} \textmd{where} \hspace{1cm} u_{i} = V^{T} \tanh(W_{e}e_{i}+b) 
\end{equation}
The matrix $W_{e}\in \mathbb{R}^{A\times H}$ and the bias vector $b \in \mathbb{R}^{A\times 1}$ are learned in a dense layer. The learned vector $V$ combines the atom annotations from the attention space, through a dot product the output of which is fed to a softmax layer, to obtain the SMILES attention weights.\\
Alternatively, we devised a contextual attention (CA) mechanism that utilizes the gene expression subset $G$ as context in determining the attention weights $\alpha_{i}$ according to the following equation (see \autoref{fig:attention_mechanism}\textcolor{myblue}{D}):
\begin{equation}
\label{eq:contex_att}
u_{i} = V^{T} \tanh(W_{e}e_{i} + W_{g}G) \hspace{1cm} \textmd{where} \hspace{1cm}  W_{g} \in \mathbb{R}^{A\times |G|}
\end{equation}
First, the matrices $W_{g}$ and $W_{e}$ project both genes $G$ and the embedded sequence element $e_{i}$ into the attention space, $A$.
Adding the gene context vector to the projected token ultimately yields an $\alpha_{i}$ that takes into account the relevance of a compound substructure given a gene subset $G$.

In both attention mechanisms, the encoded smiles are obtained by filtering the inputs with the attention weights. However, operating directly on the embeddings, the attention-based encoder disregards positional information whilst exclusively exploring information of individual tokens (i.e., atoms and bonds). In an attempt to combine atomic and regional, submolecular interactions, we tested a (shallow) multichannel convolutional attentive encoder (MCA) with two parallel kernel sets of sizes [$H$, 5] and [$H$, 11]. The resulting feature maps were fed to 2 separate contextual attention layers; a third one operated directly on the SMILES embedding so as to incorporate information from individual tokens. For all models, the output of the smiles encoder was concatenated with the output of the gene expression encoder before the regression was completed by a set of feedforward layers.

\subsection{Model evaluation}
\label{sseq:data_split}

\begin{figure}[!hbt]
\centering
\includegraphics[width=.7\linewidth,trim={0 8.8cm 0 0.5cm},clip]{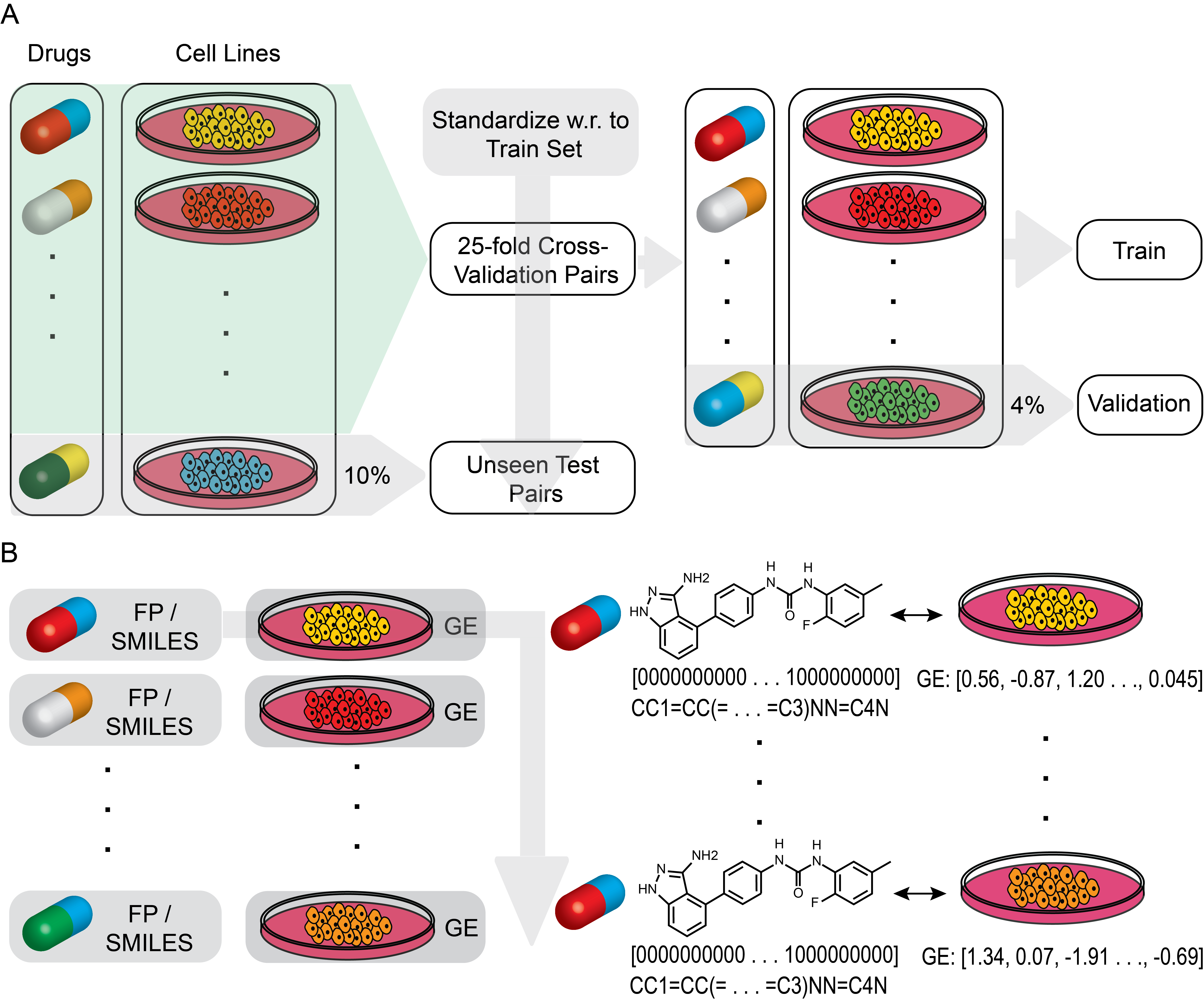}%
\caption{\textbf{Data preparation, split and model evaluation strategy.} We adopted a strict train-validation-test approach, ensuring that none of the cells and drugs in the test or validation set were seen by the trained model.  All drugs and pan-cancer cell lines within the GDSC dataset were lined up and a 10\% random subset of both drugs and cell lines were separated as test dataset. The remaining 90\% of the drugs and cell lines were used for model training and evaluation in a 25-fold cross-validation scheme.}
\label{fig:split}
\end{figure}

We adopted a strict data split approach ensuring no cell line profile or compound structure within the validation or test set has been seen by our models prior to validation or testing. 10\% subsets of the total number of 208 compounds and 985 cell lines from the GDSC database were set aside to be used as an unseen test dataset to evaluate the trained models (as shown in \hyperref[fig:split]{Figure 4A}). 
The remaining 90\% of compounds and cell lines were then used in a 25-fold cross-validation scheme for model training and validation. In each fold, 4\% of the drugs and 4\% of cell lines were separated as validation dataset and the remaining drugs and cell lines were paired and fed to the model for training, as shown in \hyperref[fig:split]{Figure 2A}. Each drug in a subset (train, validation, test) was paired with gene expression profile (GEP) in the subset that the drug had been screened with. All models ingested the data as either FP-GEP or SMILES-GEP pairs and returned normalized IC50 drug sensitivity. 
\subsection{Training procedure}
\label{sseq:training}
All described architectures were implemented in \texttt{TensorFlow 1.10}\footnote{\url{https://github.com/tensorflow/tensorflow}} and optimized MSE with Adam ($\beta_{1}=0.9$, $\beta_{2}=0.999$, $\varepsilon=1\mathrm{e}{-8}$) and a decreasing learning rate~\cite{kingma2014adam}. $H$=16 for all SMILES encoder and $A$=256 for all but the convolutional encoder (64). In the final dense layers of all models we employed dropout ($p_{drop}=0.5$), batch normalization and a sigmoidal activation.
All models were trained with a batch size of 2,048 for a maximum of 500k steps on a cluster equipped with \texttt{POWER8} processors and a single \texttt{NVIDIA Tesla P100}.

\section{Results}
\label{seq:results}

\subsection{Performance comparison}
\autoref{tab:performance} compares the test performance of all six models trained using a 25-fold cross validation scheme. 
\begin{table}[!htb]
\centering
\caption{
\textbf{Performance of the considered architectures on test data}.
Results for all the architectures trained during the 25-fold cross validation process. The first column reports the best result in terms of RMSE between predicted and true IC50 values on test data, whereas the second column shows the average across all 25 models. Interestingly, attention-only models statistically significantly outperform all others, including models trained on fingerprints (star indicating a significance of $p$<0.01 compared to the baseline encoder). 
}
\label{tab:performance}
\scalebox{0.9}{
\begin{tabular}{ccccccc}
\toprule \toprule
\multirow{2}{*}{\bfseries Encoder type} & \multirow{2}{*}{\bfseries Drug structure} &
 \multicolumn{2}{c}{\bfseries RMSE} \\ 
& & Best & Average &  \\  \midrule 
\textbf{Deep baseline (DNN)} & Fingerprints & 0.114  &  0.123 $\pm$ 0.008 \\ \midrule
\textbf{Bidirectional recurrent (bRNN)} & SMILES & 0.106 &  0.118 $\pm$ 0.007 \\ \midrule
\textbf{Stacked convolutional (SCNN)} & SMILES & 0.120 &  0.133 $\pm$ 0.012 \\ \midrule
\textbf{Self-attention (SA)} & SMILES & \bfseries0.089  & 0.112* $\pm$ 0.007 \\ \midrule
\textbf{Contextual attention (CA)} & SMILES & 0.095 & \bfseries0.110* $\pm$ 0.008 \\ \midrule
\textbf{Multichannel convolutional attentive (MCA)} & SMILES & 0.106 & 0.120 $\pm$ 0.001 \\ \midrule 
\end{tabular}}

\end{table}
The contextual attention models yielded on average the best performance in predicting drug sensitivity (IC50) of unseen drug-cell pairs. As IC50 was normalized to [0,1], the RMSE for the CA model implies an average deviation of 11\% of the predicted IC50 values from the true values. 

As the most prevalent technique to encode sequences, we found the bRNN to match but not surpass the performance of the baseline model. The SCNN, whose encoded feature maps consider information from even longer spatial segments of the SMILES than the bRNN (due to the stacking), performed significantly worse than the baseline, as assessed by a one-sided Mann-Whitney-U test ($U$=126, $p$<2e-4).

We therefore hypothesized the atomic features on the SMILES-token level (rather than submolecular interactions) to be most predictive of a drug's efficacy (such as mere atom counts). Utilizing attention-based models that operated directly on the SMILES embedding, we extracted two models (SA, CA) that were superior to all others (e.g., CA vs. DNN: $U$=42, $p$<9e-8, SA vs. DNN: $U$=82, $p$<5e-6). Incorporating genomic information into the attention process (CA vs. SA) had not a \textit{signifcant} effect on prediction performance. Surprisingly, neither complementing the SMILES embedding with positional encodings (like in \citet{vaswani2017attention}) nor adding the attention mechanism after the recurrent encoder was found to be beneficial for the presented models during the optimization process on the cross-validation folds. Ultimately, the MCA model aimed towards complementing the token-level information (that was provably benefical for the attention-only models), with spatially more holistic chemical features within the same model. However, the MCA is found to perform comparable to the bRNN, whilst not reaching the attention-only models. Overall, these results suggest that token-level (atoms, bonds) but not subregional information is most predictive for drug sensitivity. 
\subsection{Attention analysis}

Considering the best SA model trained (RMSE=0.089 and Pearson correlation $\rho$=0.66 on the test data), we analyzed its predictions for a selection of drugs and cell lines included in the test set (see~\autoref{tab:gene_attention}).
\begin{table}[!htb]
\centering
\caption{
\textbf{IC50 prediction for a panel of ten cell-drug pairs }.
Listed in the table are the predicted and true IC50 values for a selection of 10 unseen drug-cell pairs within the test dataset. In addition, the top-5 attented genes given by GAE are listed for each cell line.}
\label{tab:gene_attention}
\scalebox{0.74}{
\begin{tabular}{ccccccc}
\toprule
\multirow{2}{*}{\bfseries Drug} & \multirow{2}{*}{\bfseries Cell line} &  \multirow{2}{*}{\bfseries Cancer type} &  \multirow{2}{*}{\bfseries Top-5 attended genes} &  \multicolumn{2}{c}{\bfseries IC50} \\
 &  &  &  & Predicted &  Measured \\  \midrule
\bfseries Afatinib &	UMC-11	& lung (NSCLC) &	F13A1, MYH4, ATOH8, SEMA4A, NES	& 0.505	& 0.493 \\ \midrule
\bfseries BX-912 &	YH-13 & 	glioma	& RNASE2, HOXA13, CBR3, FABP1, HDC &	0.532 &	0.5 \\ \midrule
\bfseries GSK319347A	& EW-12	& bone	& CD300A, RHBDL2, NES, TFF3, SOCS1	& 0.597 & 	0.7 \\ \midrule
\bfseries JW-7-24-1	& OVTOKO &  	ovary	& HDC, EIF2A, RNASE2, ANGPTL6, CBR3 & 	0.502	& 0.49 \\ \midrule
\bfseries PI-103 &	MV-4-11	& leukemia &	TFF3, ATOH8, RBP2, ITIH3, GRIP1	& 0.362	& 0.33 \\ \midrule
\bfseries TGX221 &	SW962	& urogenital system	& CBR3, RNASE2, FABP1, HDC, SH3D21	& 0.621 &	0.66 \\ \midrule
\bfseries S-Trityl-L-cysteine & 	NCI-H187 &	lung (SCLC) &	RHBDL2, NR1H4, MYH4, NES, APCS	& 0.535 & 0.502\\ \midrule
\bfseries Fedratinib &	BL-41 &	lymphoma &	TFF3, ATOH8, RBP2, MAPK7, ARHGEF33	& 0.382	& 0.428 \\ \midrule
\bfseries Tipifarnib &	RCC10RGB & 	kidney	& EIF2A, HDC, CBR3, PIK3R5, HOXA13 &	0.542	& 0.544 \\ \midrule
\bfseries Midostaurin &	GAK &	skin	& SVOP, FABP1, HDC, F13A1, FGFR3 &	0.507 &	0.477 \\ \midrule
\end{tabular}}

\end{table}
The genes highlighted by the computed attention weights showed a significant enrichment~\cite{kuleshov2016enrichr,ashburner2000gene} of the JAK-STAT signaling pathway, a known therapeutic target in cancer~\cite{o2015jak}.

\begin{figure}[!htb]
\centering
\includegraphics[width=1\linewidth]{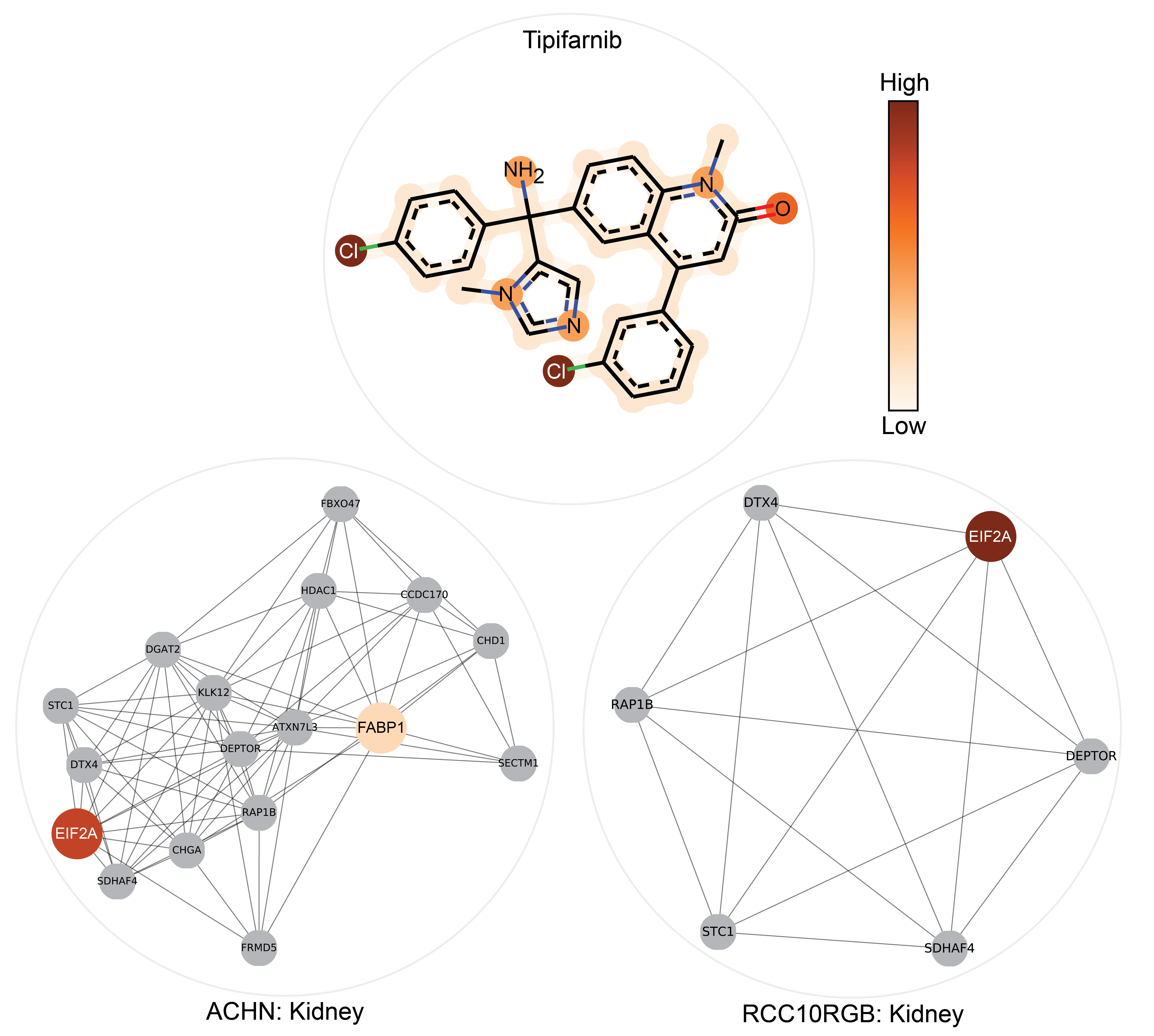}%
\caption{
\textbf{
Attention weights on genes and drugs.
}
We report here attention weights for Tipifarnib (top) and two kidney cell lines: ACHN (bottom left) and RCC10RGB (bottom right).
The attention weights are computed using the best SA model with their intensity being encoded using an orange color map (light to dark).
In Tipifarnib we highlight elements of the drug molecular structure using the learned weights.
In the two cell lines we highlight genes with highest attention weights while displaying their neighborhood in the STRING networks.
}
\label{fig:attention}
\end{figure}

As a case study, we chose an anticancer compound from~\autoref{tab:gene_attention}, Tipifarnib, and studied the SMILES attention weights given by the SA model. The attention weights for Tipifarnib are visualized in~\autoref{fig:attention}. As shown in figure 5, rudimentary or frequent atoms such as C or H are given a negligible weight while chlorine is given the highest weight followed by the amid group, -NH2, O, N and the covalent bonds. As an additional case study, at the bottom of~\autoref{fig:attention}, the STRING protein neighborhoods for top-weighted genes of two kidney cancer cell-lines are illustrated. Interestingly, even if the cell lines are not genomically identical, both computed attention weights underline the role of EIF2A, a key gene for tumor initiation~\cite{sendoel2017translation}.

\section{Discussion}
\label{sec:discussion}
In this work, we presented a multi-modal approach to neural prediction of drug sensitivity based on a combination of 1) molecular structure of drug compounds 2) the gene expression profile of cancer cells and 3) intracellular interactions incorporated into a PPI network. We adopted a strict model evaluation approach in which we ensured that no cells or compounds in the validation or test datasets are ever seen by the trained model. Despite our strict evaluation strategy, our best model (CA) achieved an average standard deviation of 0.11 in predicting normalized IC50 values for unseen drug-cell pairs. We demonstrated that using the raw SMILES string of drug compounds in combination with an array of various encoders, we were able to surpass the predictive performance reached by utilizing Morgan fingerprints (the baseline model). Notably, encoders based solely on attention mechanisms (i.e., SA and CA) exhibited superior performance compared to RNN or CNN encoders, indicating that drug sensitivity may be more predictable from molecular information on an atomic level rather than longer range dependencies. This observation however, must be further investigated and verified as a future direction. The atom-level attention weights obtained in our models may be used to identify atomic features or bonds that are deemed more informative by the algorithm. We illustrated the atom-level attention weights returned by our SA model for Tipifarnib and highlighted the atoms and bonds that were given a higher weight.   

In addition, we devised and utilized a gene attention encoder that returned attention weights for genetic profile of cells and thereby enabled us to better interpret the results and identify genes that were most informative for IC50 prediction. As a testament to the validity of the gene attention weights, we observed that different cell lines of the same organ tend to have similar (but not the same) gene attention weights. We envision that the approach used in our attention-based gene encoder will be of great utility in deep learning applications where a  high level of interpretability is desired. As a future direction, it would certainly be beneficial to include gene expression profiles from healthy cell lines in the training samples so as to ensure that the model extracts features that can differentiate between healthy and cancerous cells. 

Due to the multi-modal nature of our models, the interpretability offered by the attention mechanism and the stringent evaluation criteria employed in training and evaluating our models, we envision our models to generalize well and be applicable in drug discovery or personalized medicine where estimates on efficacy of candidate compounds on a wide range of cancer cells are required.

\bibliographystyle{unsrtnat}
\begin{footnotesize}
\bibliography{main}
\end{footnotesize}

\end{document}